\documentclass[letterpaper, 10 pt, journal, twoside]{IEEEtran}
\usepackage[english]{babel}
\usepackage{amsmath} 
\usepackage{amssymb}  
\usepackage{graphicx}
\usepackage{subfigure}
\usepackage{multirow}
\usepackage{amsfonts}
\usepackage{hyperref}
\usepackage{tabularx}
\usepackage[noend]{algpseudocode}
\usepackage{algorithm}
\usepackage{bm}
\usepackage{enumerate}
\usepackage{verbatim}
\usepackage{float}
\usepackage{siunitx}
\usepackage{mdframed}
\usepackage{adjustbox}
\usepackage{cleveref}
\usepackage{authblk}
\usepackage{color}
\usepackage{soul}
\usepackage{cite}
\usepackage{booktabs}
\usepackage{amssymb} 
\usepackage{pifont} 

\usepackage{xcolor}
\definecolor{navy}{RGB}{0,0,128}
\definecolor{myPurple}{RGB}{150, 130, 205}
\definecolor{myYellow}{RGB}{204, 170, 0} 

\newcommand{\cmark}{\textcolor{green}{\checkmark}} 
\newcommand{\xmark}{\ding{55}} 

\ifCLASSINFOpdf
\else
\fi

\begin{document}

\title{Audio-Visual Contact Classification for Tree Structures in Agriculture
}

\author{
  \IEEEauthorblockN{Ryan Spears\textsuperscript{\textasteriskcentered}, Moonyoung Lee\textsuperscript{\textasteriskcentered}, George Kantor, Oliver Kroemer} \\
  \IEEEauthorblockA{
    Carnegie Mellon University \\
    \{rspears, moonyoul, gkantor, okroemer\}@andrew.cmu.edu
  }
  \thanks{\textsuperscript{\textasteriskcentered}\,Denotes co-lead authors (equal contribution).}
}


\maketitle

\begin{abstract}
Contact-rich manipulation tasks in agriculture, such as pruning and harvesting, require robots to physically interact with tree structures to maneuver through cluttered foliage. Identifying whether the robot is contacting rigid or soft materials is critical for the downstream manipulation policy to be safe, yet vision alone is often insufficient due to occlusion and limited viewpoints in this unstructured environment. To address this, we propose a multi-modal classification framework that fuses vibrotactile (audio) and visual inputs to identify the contact class: leaf, twig, trunk, or ambient. Our key insight is that contact-induced vibrations carry material-specific signals, making audio effective for detecting contact events and distinguishing material types, while visual features add complementary semantic cues that support more fine-grained classification. We collect training data using a hand-held sensor probe and demonstrate zero-shot generalization to a robot-mounted probe embodiment, achieving an F1 score of 0.82. These results underscore the potential of audio-visual learning for manipulation in unstructured, contact-rich environments. Website and source code is available at  {\url{https://tree-classification.vercel.app/}}
\end{abstract}


\section{Introduction}

Humans are remarkably adept at maneuvering their arms through cluttered and unstructured spaces by intuitively feeling their way through the environment. This capability allows us to perform complex tasks with minimal visibility, relying on rich multisensory feedback rather than visual information alone. Consider, for example, the task of picking an apple where the scene is heavily occluded by leaves and branches. Despite this clutter, humans exhibit an extraordinary ability to feel their surroundings and maneuver their arms. We can push aside small branches, ignore harmless foliage, and reroute arm motions around obstacles like trunks, all without full visibility of the scene. These actions are enabled by innate multisensory integration of vision and haptics-guided adaptive interaction~\cite{roy2021machine}.
For robots to achieve contact-rich manipulation skills, such as pruning or harvesting, in real-world agricultural settings requires more than visual perception to accurately perceive and react to contact events. This need arises because much of the interactions with the environment  occurs at close range, outside the camera's field of view, or under heavy occlusion that vision cannot observe. To operate in such environments, robots can utilize haptic signals to reason about safe versus unsafe interactions in cluttered conditions. The ability to classify contacted objects (e.g., distinguishing brushable leaves from rigid branches) can support semantic and spatial reasoning for the downstream reactive planner or controller, such as deciding whether or not to keep pushing on a contact surface when maneuvering through dense foliage. 

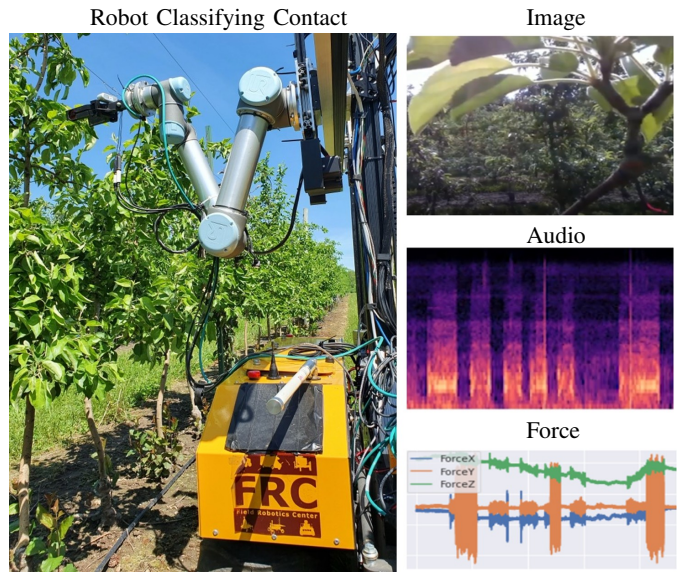
\begin{figure}[!t]
    \centering
    \includegraphics[width=1\columnwidth]{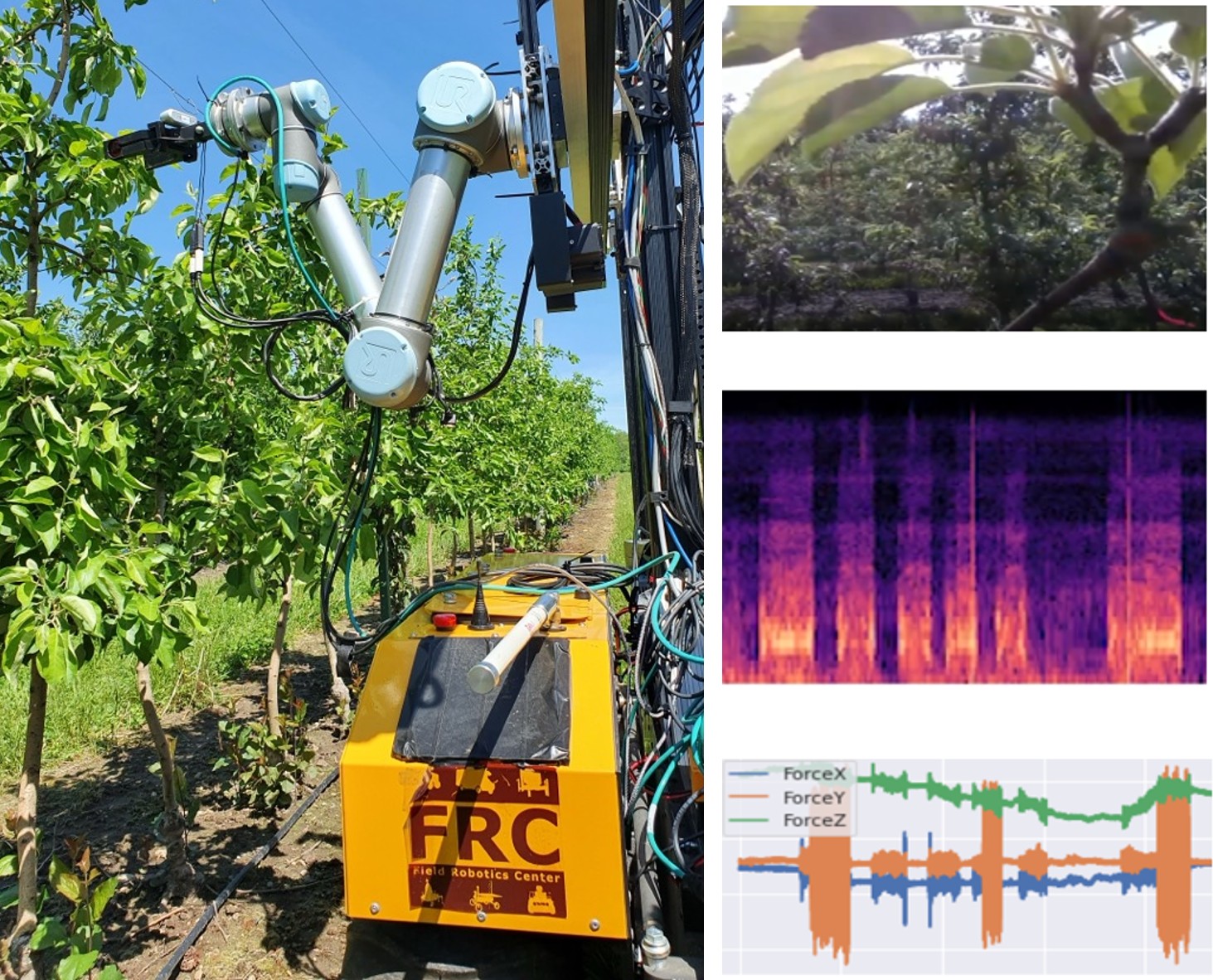}
    \setlength{\unitlength}{1cm}
    \begin{picture}(0,0)
    \put(2.5, 2.2){\textcolor{black}{\small Force}}
    \put(2.5, 4.8){\textcolor{black}{\small Audio}}
    \put(2.5, 7.7){\textcolor{black}{\small Image}}
    \put(-3.3, 7.7){\textcolor{black}{\small Robot Classifying Contact}}
    \end{picture}
    \vspace{-10pt}
    \caption{Robotic arm in an apple orchard. To distinguish safe versus unsafe contacts in a cluttered environment, robot uses multisensory inputs to classify contact events.}
    \label{fig:title_photo}
\end{figure}
Despite the benefits of utilizing multi-sensory input to guide physical interactions, it remains an open-research problem about which tactile sensing mechanisms to adopt and how to represent tactile features useful for manipulation. Contact sensors that operate in cluttered environments needs to cover a large sensing area and be robust to wear-and-tear~\cite{lee2024sonicboom} . Hence, scaling up haptic sensors that operate from direct contact regions is not trivial, as shown with existing haptic modalities, such as magnetic-based ~\cite{hellebrekers2019soft}, camera-based elastomers~\cite{yuan2017gelsight}, and pressure-sensitive robot skin~\cite{bhirangi2021reskin}. Motivated by these constraints, we explore the use of contact microphones, which can indirectly detect vibrotactile signals from contact events and cover larger areas due to vibration propagation through material. Importantly, they are safer to deploy, as they can be embedded within surface layers, reducing exposure to abrasion from repeated interaction—unlike exposed direct contact sensors.

Despite their promise, contact microphones pose unique challenges. Unlike traditional microphones that \textit{passively} capture ambient audio in the air, contact microphones only detect signals during \textit{active} interaction, making the data acquisition process more difficult. Collecting diverse audio samples requires varied and deliberate contact interactions, which are non-trivial to design or execute with robotic exploration. While many areas of robotics have benefited from large-scale pretraining, applying such strategies to contact microphones is not straightforward. Contact audio from interactions with tree structures (e.g., bristling, scratches, and thumps) differ significantly from the types of audio found in internet-scale datasets (e.g., speech or ambient noise), thus limiting the effectiveness of conventional pretrained models for this domain. ~\cite{thankaraj2023sounds, liu2024maniwav}. 
To address this, we introduce a framework that simplifies data collection and supports transfer to robotic deployment. Our approach uses a hand-held multisensory probe for a human operator to efficiently gather data from natural interactions in the field. Contact audio signals acquired through the hand-held probe differ from robot-mounted sensor due to the additional noise added from the mobile robot's  diesel power generator and  robot arm's motors. We therefore leverage audio preprocessing and pretraining techniques to enable zero-shot transfer across embodiments. Combining strengths of both audio and visual modalities, our framework achieves higher contact detection and classification results than the baseline of using a single modality.
Reasoning about contacts is an essential capability for deploying robots in contact-rich agricultural environments, where physical interactions are part of the task. While our current system does not yet close the loop between classification and downstream motion planning, it marks an important step toward such multisensory action pipelines. We also demonstrate that audio-visual model trained on human hand-held probe can zero-shot transfer to robot-probe with aid of internet-based pretraining. This work lays the groundwork for future in-the-loop systems where robots use contact reasoning to act safely and adaptively in agricultural settings.

Our contributions are thus: (1) we design robust audio feature representations and leverage internet-scale pretraining to enable zero-shot transfer across hardware embodiments; (2) we provide empirical insights into the complementary strengths of audio and vision -- where audio excels at detecting contact events and vision at contact classification; and (3) we release an open-source multisensory dataset for contact-rich interactions in agricultural environments.

\section{Related Works}

\textbf{Contact-Rich Manipulation in Agriculture:}
Robotic manipulation in agriculture is particularly challenging due to perceptual occlusion\cite{kim2023occlusion} and trees' deformation during contact, which can deviate from the robot’s internal world model\cite{lagrassa2024task}. Traditional approaches follow a sense-plan-act pipeline, often using vision-based 3D reconstruction of plants~\cite{Kantor_visual, sorghum_3d, kim2023occlusion}. However, even with occlusion-aware perception, downstream planning remains difficult. To overcome this, recent works have explored learning visuomotor policies in simulation via supervised\cite{jacob2024learning, kim2023robotic} or reinforcement learning\cite{jacobgentle}. Given the sim-to-real gap in modeling interactions with deformable objects like plants, instead of learning an accurate model, other approaches leverage real-time haptic feedback to adopt the manipulation policy. Our work builds on this direction, where similar works have used contact microphones to infer semantic and spatial cues in unstructured, contact-rich environments~\cite{lee2024sonicboom, liu2024sonicsense, yi2024visual}.

\textbf{Multisensory Classification with Audio:}
With the rise of multimodal architectures, combining vision and haptics has become increasingly relevant. Audio signals have been used for adaptive policy learning\cite{du2022play, clarke2018learning},  dynamics parameter mapping\cite{thankaraj2023sounds}, and inverse model learning\cite{gandhi2020swoosh}. Our work focuses on classification using passive contact microphones, which require physical interaction to generate signal. Related efforts have explored object classification via rattling objects inside a box\cite{gandhi2020swoosh, chen2021boombox} or food category classification\cite{IAM_cutting, IAM_food}. Other studies employ additional hardware with sound emitter and receiver for active acoustic sensing for contact region or state estimation\cite{brock2023_journal, yoo2024poe, SamsungAI_2023sonicfinger}, though these are limited to smaller sensing regions like between the robot fingertips.

\textbf{Audio Pretraining for Manipulation:}
Recent work in self-supervised and contrastive learning has enabled large-scale audio-visual pretraining\cite{yang2024binding}, but its effectiveness on contact microphone data remains unclear due to domain mismatch ~\cite{thankaraj2023sounds}. Prior studies show that models trained from scratch often outperform those using pretrained features\cite{thankaraj2023sounds, liu2024maniwav}. In line with~\cite{mejia2024hearing}, we find that with proper preprocessing, audio pretraining can still improve performance even in this specialized setting. %

\section{Data Collection}

\subsection{Sensor Probe for Contact Interaction}

\begin{figure}[t]
    \centering
    \includegraphics[width=0.9\columnwidth]{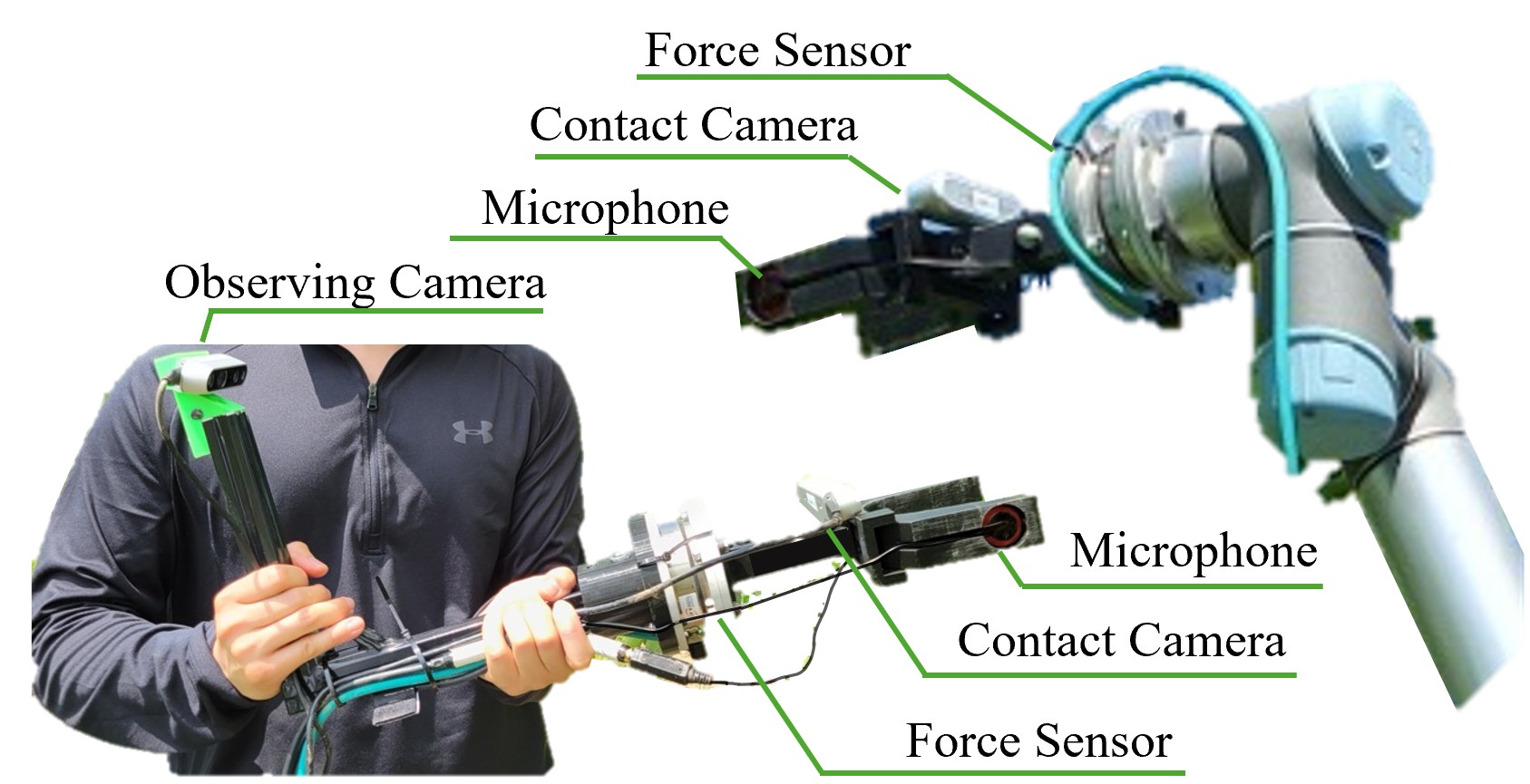}
    \caption{Multi-modal sensor suite in two different embodiments. (Left) Hand-held to facilitate data collection (Right) for robot deployment.}
    \label{fig:sensor_suite}
    \vspace{-10pt}
\end{figure}

\begin{figure}[t]
    \centering
    \includegraphics[width=1\columnwidth]{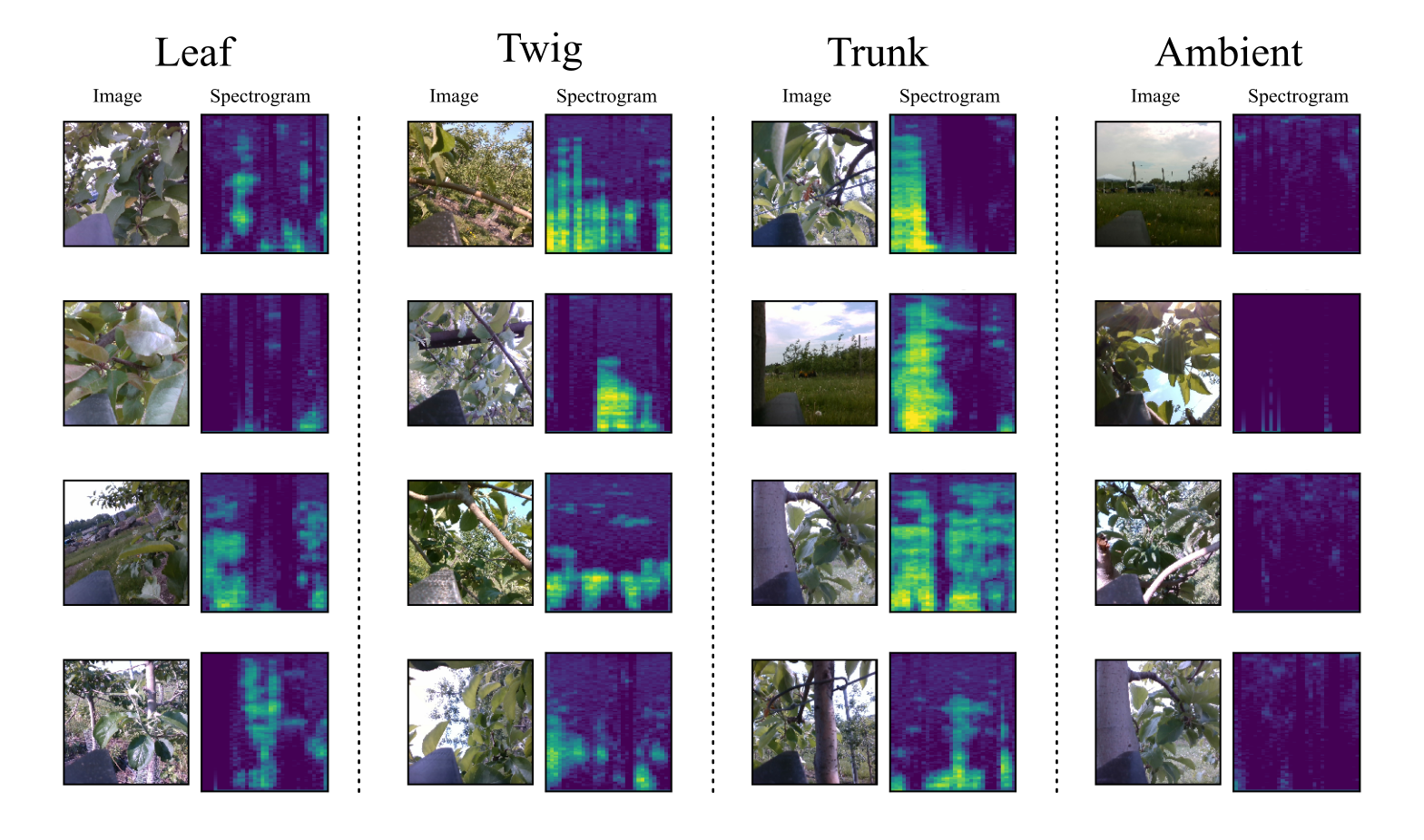}
    \vspace{-20pt}
    \caption{Audio-image pair of four contact classes. In both sensing modalities, there are distinct patterns between each category that a model can learn from both the spectrograms and images to assist with classification. }
    \label{fig:experiments}
\end{figure}

To address the practical challenges of collecting extensive contact data directly with the robot in field environments which is time-consuming, dangerous, and resource-intensive, we developed a hand-held sensor probe to enable an easier, scalable data collection process in the field. The end-effector of our sensor probe is designed to resemble a parallel-jaw gripper system commonly used in existing apple harvesting robots~\cite{yu2021lab}. As shown in Fig.~\ref{fig:sensor_suite}, the same sensor suite is mounted on the robot end-effector and on the hand-held probe for an identical sensing configuration across embodiments. The flexibility of the hand-held device allows human operators to generate a broad range of contact interactions with variations in motion trajectory, velocity, and duration.

The sensor suite consists of three complementary modalities: (1) an Intel RealSense D435 camera (a front-facing contact-view camera), (2) a contact microphone for capturing vibrotactile signals, and (3) an ATI FT24252 force-torque sensor mounted at the wrist. The contact microphone is a low-cost piezoelectric sensor with high impedance and weak signal strength, which requires amplifying the input signal. We use a UMC404HD audio interface with 24-bit/192kHz converters and manually tune the gain to prevent clipping while maintaining sensitivity to subtle contact events, such as the rustling of leaves. Although we do not use force modality for our classification, we utilize it for verifying our automatic data annotation, as discussed in Sec. \ref{sec:annotation}. All sensors are synchronized using ROS, with sampling rates of 22kHz for audio, 250Hz for force-torque, and 30Hz for the camera.
\vspace{-10pt}

\subsection{Data Collection Procedure}

Our dataset was collected at the University of Massachusetts Amherst’s Cold Spring Orchard across three apple varieties: Fuji, Gala, and Honeycrisp. In total, we recorded 300 ROS bag files (250 with a hand-held probe; 50 with a robot). Each bag contains two contact interactions with a single object class over approximately 15 seconds. To facilitate automated labeling, each interaction is constrained to a single class (e.g., leaf, twig, trunk). The contact motion involves pushing into the target object and pulling out, similar to a smooth prodding motion.

For the robot trials, the sensor suite was attached to a 7-DoF UR5 robot arm mounted on a mobile platform~\cite{silwal2022bumblebee}. The robot base was teleoperated to drive infront of a desired tree, the robot arm manually positioned to face a target object using gravity-compensation mode. From there, the arm executed a predefined linear trajectory under velocity control to push into and pull away from the object. For data collected with the hand-held probe, prodding motions were also used.

\begin{figure}[t]
    \centering
    \includegraphics[width=1.0\columnwidth]{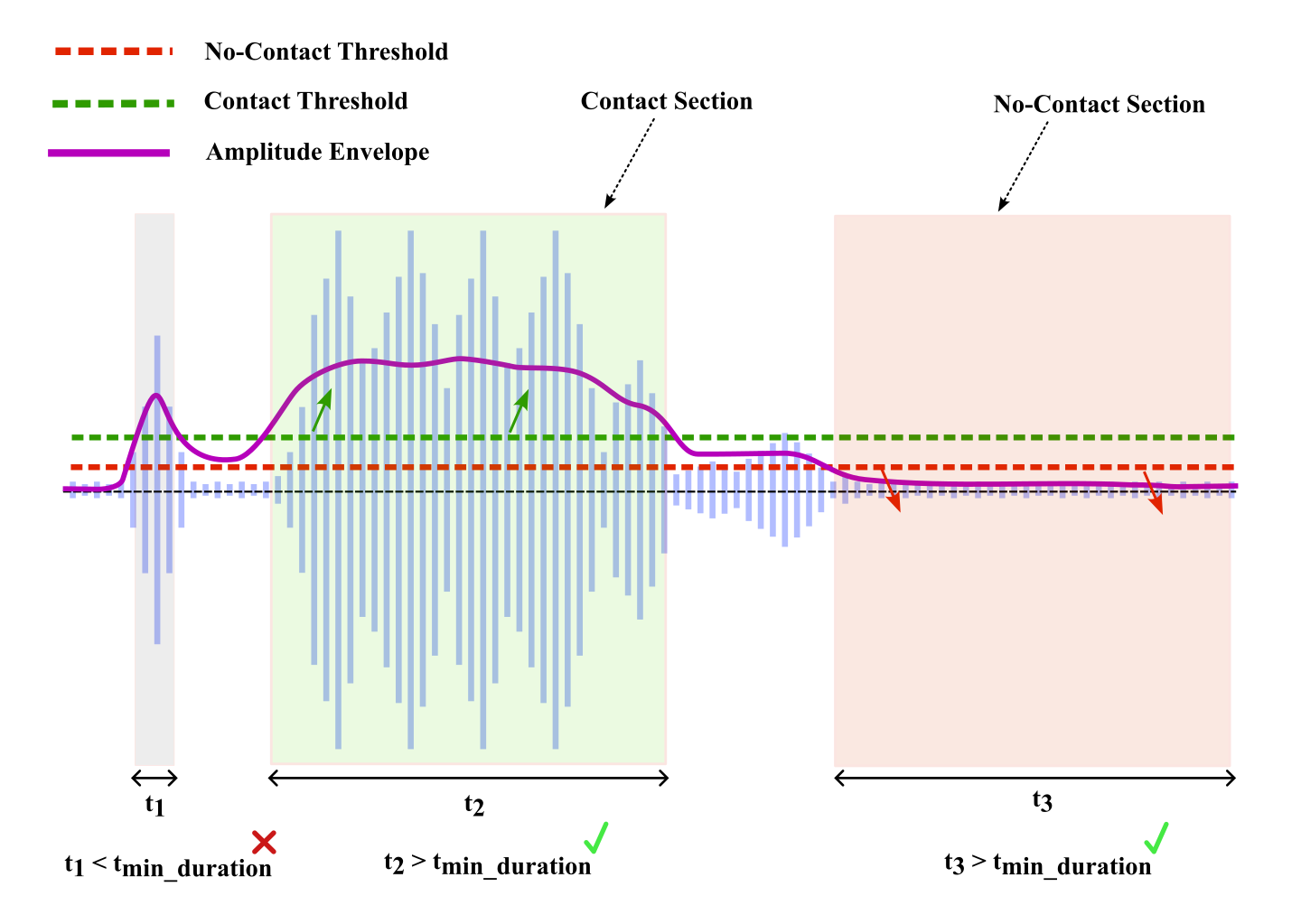}
    \vspace{-20pt}
    \caption{Audio-visual dataset automatically annotated into contact (green) and non-contact (red) intervals using audio-based segmentation. Contact segments are identified by applying a dynamic threshold to the smoothed audio amplitude; values above the threshold are labeled as contact, and those below as non-contact.}
    \label{fig:gui}
    
\end{figure}

\section{Data Processing}
Given the sensor hardware and data collection procedure, we now describe how the raw data is curated for training. The process involves annotating audio and image streams with semantic and temporal labels. To avoid tedious manual labeling, we automate the detection of relevant contact segments. Since each trial involves interaction with a single object class, annotation simplifies to identifying periods of contact with the corresponding class versus ambient background.

\subsection{Data Annotation Automation} ~\label{sec:annotation}

To efficiently analyze large sensory datasets, we developed an automatic segmentation pipeline that separates contact from non-contact intervals using audio signals. This process is critical for labeling and training. We compute a smoothed moving average of audio amplitudes and dynamically determine a thresholds per sample that distinguishes background noise from active contact events. To ensure meaningful interactions, we enforce a minimum segment duration of one second and merge neighboring segments that are closely spaced in time. Notably, contact segments may contain brief silences, but for a region to be labeled as no-contact, the average signal must remain entirely below the threshold. Our segmentation technique, as shown in Figure \ref{fig:gui}, can be expressed with the following equations:

\interdisplaylinepenalty=2500
\begin{small}
\begin{IEEEeqnarray}{rCl}
T_{\mathrm{contact}} &=& F_{\mathrm{noise}}
   + (F_{\mathrm{signal}} - F_{\mathrm{noise}})\,\alpha_{\mathrm{offset}},\nonumber\\
T_{\mathrm{non\!-\!contact}} &=& \beta_{\mathrm{factor}}\,T_{\mathrm{contact}}
\label{eq:thresholds}\\[6pt]
C(t) &=&
\begin{cases}
1, & E(t) > T_{\mathrm{contact}},\\
0, & E(t) \le T_{\mathrm{contact}},
\end{cases}
\label{eq:classification}\\[6pt]
S_{\mathrm{final}} &=& \{[t_{\mathrm{start}}^i,\,t_{\mathrm{end}}^i]\;\mid\nonumber\\
&&\quad C(t)=c\;\forall\,t\in[t_{\mathrm{start}}^i,t_{\mathrm{end}}^i],\nonumber\\
&&\quad t_{\mathrm{end}}^i - t_{\mathrm{start}}^i \ge \delta_{\min},\nonumber\\
&&\quad t_{\mathrm{start}}^{i+1} - t_{\mathrm{end}}^i \le \gamma_{\mathrm{squeeze}}\}
\label{eq:segmentation}
\end{IEEEeqnarray}
where $T_{\mathrm{contact}}$ is the contact amplitude threshold, computed as the noise floor plus an offset fraction of the signal peak; 
$F_{\mathrm{noise}}$ is the 10th percentile of the smoothed average signal (noise floor); 
$F_{\mathrm{signal}}$ is the 90th percentile of the smoothed signal (signal peak); 
$\alpha_{\mathrm{offset}}$ is the contact-threshold offset parameter; 
$T_{\mathrm{non\!-\!contact}}$ is the non-contact threshold, equal to $\beta_{\mathrm{factor}}\,T_{\mathrm{contact}}$; 
$\beta_{\mathrm{factor}}$ is the non-contact factor; 
$C(t)\in\{0,1\}$ is the binary contact label; 
$E(t)$ is the smoothed signal value; 
$S_{\mathrm{final}}$ is the set of merged, duration-filtered intervals $[t_{\mathrm{start}}^i,t_{\mathrm{end}}^i]$; 
$\delta_{\min}$ is the minimum segment duration; 
and $\gamma_{\mathrm{squeeze}}$ is the maximum gap for merging adjacent segments.
\end{small}

Once the segmentation is complete, the resulting intervals are transferred into a visualization program. This tool overlays the computed contact intervals on the original visual recordings from the probe-mounted camera. The output is a composite video that includes a timeline bar marking contact versus no-contact intervals and frames of the original scene. Please refer to our website for further visualizations. This verification video plays a crucial role in confirming the accuracy of the segmentation process—it allows human reviewers to cross-check whether the segments correctly align with visual evidence of contact events.

We categorize contact segments into four classes: leaf, twig, trunk, and ambient. In our case, \textquotedblleft ambient" refers to no-contact sections. No-contact sections refer to an instance without significant forceful contact. By combining automatic segmentation with visual alignment, we create a rigorous, validated dataset that enhances the reliability of our learning pipeline. 

\subsection{Audio Preprocessing}

The contact microphones are specifically designed to capture vibrations through materials. While contact microphones effectively minimize the pickup of ambient audio (e.g., wind, human speech), ideally, the audio signal should only contain vibrotactile signals from contacting the tree. In our case, they contain static noise, motor noise that varies with operational load, and the robot's power generator vibrations, all of which need to be removed. We implement spectral gating \cite{sainburg2020finding} to filter out background noise using a pre-recorded reference signal of both the probe and robot data without any contact, effectively isolating only the contact signal from both embodiments as done in \cite{lee2024sonicboom}. As one key challenge is generalizing from handheld probe data to robotic data, we further apply data augmentation techniques during training—randomly adjusting pitch and noise levels—and even injecting sampled motor noise into probe recordings to more closely replicate the robot’s operating conditions. 

\subsection{Dataset Generation}
Unlike common datasets for speech recognition or sound classification, contact-based acoustic data is scarce and challenging to simulate accurately due to the physics behind energy propagation through non-uniform medium and geometric irregularities \cite{park2022biomimetic}. As such, we cannot rely on simulation to generate audio data or solely rely on audio data from the internet to train our model. 

The dataset inputs are structured as follows: each image is represented as an $244 \times 244$ RGB image with a vector shape of [3, 224, 224], while each 0.8s audio sample is transformed into a single-channel mel-spectrogram tensor of shape of [1, 128, 1024]. The training dataset comprises $\approx$7,300 samples, whereas the evaluation dataset, consists of $\approx$1,200 samples. Both datasets maintain the same structure, where each trial corresponds to a captured contact interaction lasting one second. 
 During training, an 80:20 split is employed to divide the training and validation datasets, respectively.

Our dataset $\mathcal{D}$ consists of paired audio and visual samples with contact labels, represented as  
\[
\mathcal{D} \equiv \{(A^i, I^i, y^i)\}_{i=1}^N
\]  
where $N$ is the number of samples. Each $A^i \in \mathbb{R}^{1 \times 128 \times 1024}$ denotes a Mel spectrogram extracted from contact microphone audio recorded at a 16kHz sample rate, $I^i \in \mathbb{R}^{3 \times 224 \times 224}$ represents the corresponding RGB image captured during the interaction, and $y^i \in \{\text{leaf},\ \text{twig},\ \text{trunk},\ \text{ambient}\}$ provides the ground-truth contact label as a scalar value.

In the data collection process, the recorded contact interactions were intentionally directed towards specific categories, enabling us to proactively group recordings based on the part of the tree being interacted with. Although the semantic label is fixed, \textit{when} the contact occurs during the trial period is unknown, which led us to implement a method for automatically segmenting the data into contact and no-contact sections as shown in Figure \ref{fig:gui}.

\subsection{Robot Deployment}
To demonstrate inferencing, we load the entire video of contact interactions into memory and segment it into fixed-length, overlapping windows—e.g., 30-frame clips with a 15-frame stride (50\% overlap)—then apply the same resizing, cropping, and normalization used during training. If audio is present, we extract it and convert it into mel-spectrograms (or other features) before batching each segment for a single forward pass through our multimodal model. The visual and audio embeddings are fused by the transformer to produce a contact-classification prediction, which we timestamp (using the segment’s midpoint or start frame) and either log to an output file or overlay on the video frames for visualization. A demonstration of this workflow is available on our website. %
\section{Classification Model}\label{sec:classification}

\begin{figure}[!t]
    \centering
    \includegraphics[width=1\columnwidth]{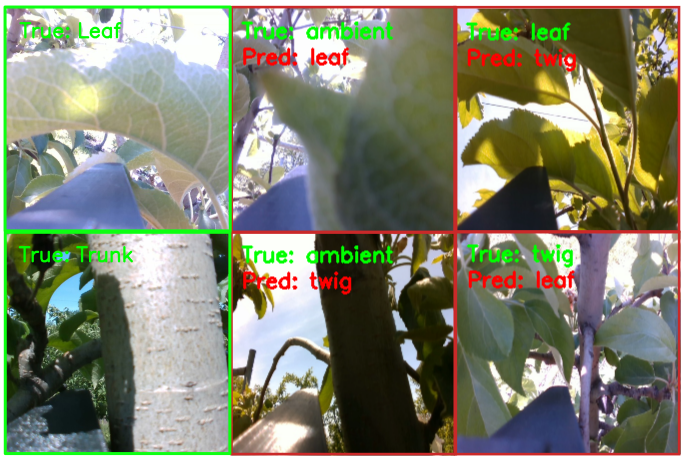}
    \vspace{-20pt}
    \caption{Examples of correct (green) and incorrect (red) predictions from the image-only model. Visual input alone struggles to distinguish subtle differences between near-contact and actual contact, as well as distinguishing ambiguous visual cues in clutter, highlighting the need for complementary haptic sensing.}
    \label{fig:confusion_image}
\end{figure}

\begin{figure*}[t]
    \centering
    \includegraphics[width=1\linewidth]{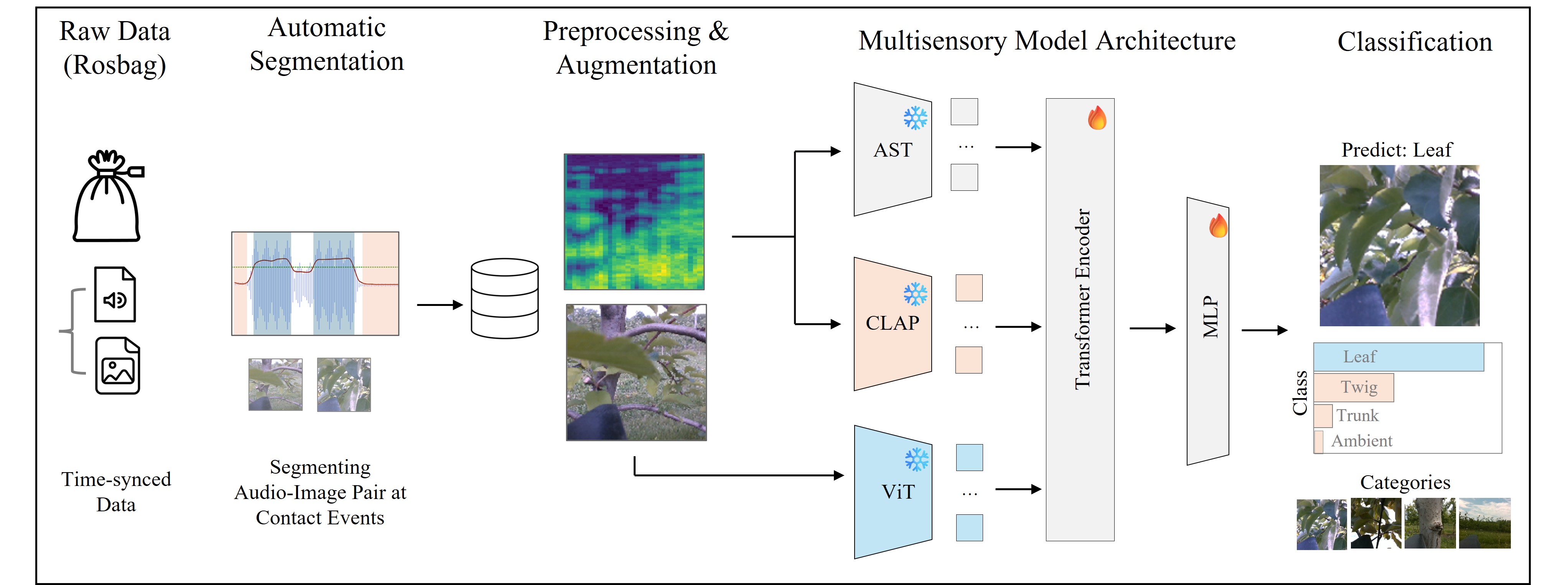}
    \vspace{-15pt}
    \caption{ Overall system diagram of our classification framework. The audio input signal is pre-processed into mel spectrograms. Each sensing modality is encoded into a latent feature before being fused by the multi-sensory self-attention transformer encoder.}
    \label{fig:sw_overview}
\end{figure*}

As shown in Figure \ref{fig:sw_overview}, we extract audio features from each 0.8s contact interaction using two complementary pretrained encoders. We select a 0.8 s window because our time-length ablation study identified it as the optimal trade-off between capturing contact dynamics and keeping inference latency low as shown in Figure \ref{fig:time}. Each 0.8-second audio clip contains 12,800 data points, given the 16kHz sampling rate, and is converted into a single mel spectrogram per sample. The Audio Spectrogram Transformer (AST) \cite{gong2021ast} processes 1024×128 mel-spectrograms as image-like inputs to generate a 768-dimensional embedding that captures fine-grained spectral and temporal patterns. In parallel, the CLAP module \cite{wu2023large}, based on the HTSAT-base encoder and pretrained on LAION-Audio-630K, projects the same spectrogram into a 512-dimensional semantic space aligned with language through contrastive learning. AST is particularly sensitive to vibrotactile signatures of contact events, while CLAP enables generalization across diverse acoustic contexts. Together, they offer low-level spectral precision and high-level semantic coherence, enhancing the model’s ability to differentiate between subtle contact categories.

Concurrently, we encode each $224\times224$ RGB frame using a ViT-B/16 pretrained on ImageNet, producing a 768-dimensional CLS token that provides complementary visual context. We concatenate the AST, CLAP, and ViT embeddings and pass them through a lightweight transformer encoder to fuse multimodal features before projecting the fused representation through an MLP classification head to obtain final logits. By keeping the three pretrained encoders frozen and training only a lightweight transformer and MLP classifier on our orchard dataset, we transfer the broad acoustic priors learned from internet-scale data to the distinctive sounds of agricultural contact, such as rustling leaves, snapping twigs, and thumping branches, rather than conventional speech or background noise.

For training, the model utilizes a cross-entropy loss function defined as:
\begin{equation}
\mathcal{L} = -\frac{1}{N} \sum_{i=1}^{N} \sum_{c=1}^{C} y_{i,c} \log(p_{i,c})
\end{equation}
\noindent
where \( N \) is the batch size, \( C \) is the number of classes, \( y_{i,c} \) is the ground-truth label for sample \( i \) and class \( c \), and \( p_{i,c} \) is the predicted probability.

We trained our model on approximately $3{,}500$ samples for up to 50~epochs using the AdamW optimizer, and a batch size of~4. Training on an NVIDIA GeForce RTX~4070 GPU with two data-loading workers took about three hours to converge. For the application of deploying this model on the robot, the inference speed of our model is roughly 14~ms to processes a one-second sample, which we can utilize a rolling-buffer with the streaming audio for live classification and robot responsiveness to contact.    %
\section{Experimental Results}\label{sec:results}

To investigate whether the proposed method of contact classification can generalize to a different embodiment, from human-held probe to robot, we train on the human hand-held data and test explicitly on robot data. All results reported below are on robot data.  For the test set collected with the robot, the audio signal is largely out of distribution from the training set due to the robot motor noise, the vibration from the power generator on the mobile platform, and the different velocity at the contact point. We quantify error  using precision, recall, and F-1 score between the predicted and the ground truth classification.

Our aim is to investigate the system through three complementary analyses that address the questions:
(\textbf{Q1}) \textit{Does audio contain sufficiently rich information to classify object contact interactions?} \textbf{(Q2)} \textit{Does combining input features (audio, vision, force) improve classification performance?}, and \textbf{(Q3)} \textit{Does pre-training improve classification performance by learning the audio/visual representation in advance? }     

\subsection{Classification Evaluation}

\begin{figure}[!t]
    \centering
    \includegraphics[width=1\columnwidth]{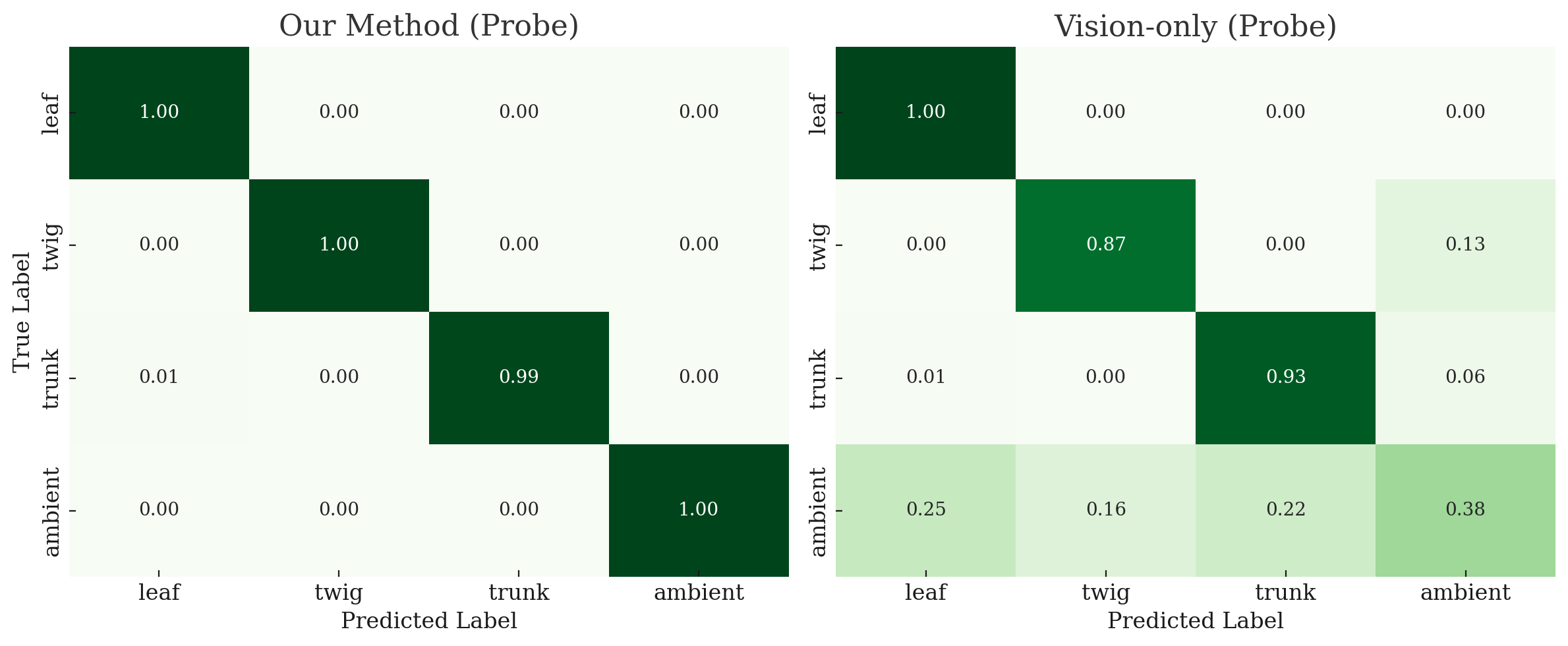}
    \caption{Probe-data confusion matrices for contact classification.  The vision-only (right) model suffers from substantial misclassifications, particularly on between ambient and contact samples while our multimodal model (left) achieves near-perfect accuracy across all four classes.}
    \label{fig:probe_confusion_matrix}
    \vspace{-10pt}
\end{figure}

\begin{table}[t]
\centering
\small
\setlength{\tabcolsep}{2pt}
\caption{Multiclass Classification Performance Comparison}
\label{tab:multiclass-performance}
\begin{tabular}{l c c c c}
\toprule
Modality                  & Pre\,-trained? & F\,-1~$\uparrow$ & Precision~$\uparrow$ & Recall~$\uparrow$ \\
\midrule
Image\,-only ViT            & \xmark       & 0.21          & 0.19          & 0.25          \\
Image\,-only ViT            & \cmark       & 0.35          & 0.45         & 0.37          \\
Audio\,-only AST            & \xmark       & 0.49          & 0.50          & 0.51          \\
Audio\,-only AST            & \cmark       & 0.52          & 0.59          & 0.56          \\
Audio\,-only CLAP           & \xmark       & 0.17          & 0.18          & 0.19          \\
Audio\,-only CLAP           & \cmark       & 0.35          & 0.34          & 0.36          \\
Audio\,-only DualAudio      & \xmark       & 0.46          & 0.52          & 0.48          \\
Audio\,-only DualAudio      & \cmark       & 0.53          & 0.50          & 0.49          \\
Audio\,-image DualAudio     & \xmark       & 0.52          & 0.51          & 0.53          \\
\textbf{Audio\,-image DualAudio} & \cmark       & \textbf{0.74} & \textbf{0.75} & \textbf{0.73} \\

\bottomrule
\end{tabular}
\end{table}

\begin{table}[t]
\centering
\small
\setlength{\tabcolsep}{2pt}
\caption{Binary Classification Performance Comparison}
\label{tab:binary-performance}
\begin{tabular}{l c c c c}
\toprule
Modality                  & Pre\,-trained? & F\,-1~$\uparrow$ & Precision~$\uparrow$ & Recall~$\uparrow$ \\
\midrule
Image\,-only ViT            & \xmark       & 0.57          & 0.58          & 0.57          \\
Image\,-only ViT            & \cmark       & 0.77          & 0.78          & 0.79          \\
Audio\,-only AST   & \xmark       & 0.86 & 0.86 & 0.85 \\
\textbf{Audio\,-only AST}            & \cmark       & \textbf{0.92}          & \textbf{0.94}          & \textbf{0.92}          \\
Audio\,-only CLAP          & \xmark       & 0.30          & 0.31          & 0.31          \\
Audio\,-only CLAP           & \cmark       & 0.58          & 0.57          & 0.59          \\
Audio\,-only DualAudio      & \xmark       & 0.83         & 0.83          & 0.83          \\
Audio\,-only DualAudio      & \cmark       & 0.80          & 0.79          & 0.81          \\
Audio\,-image DualAudio     & \xmark       & 0.89          & 0.88          & 0.90          \\
Audio\,-image DualAudio     & \cmark       & 0.90          & 0.89          & 0.91          \\
\bottomrule
\end{tabular}
\end{table}

\begin{figure}[!t]
    \centering
    \includegraphics[width=1\columnwidth]{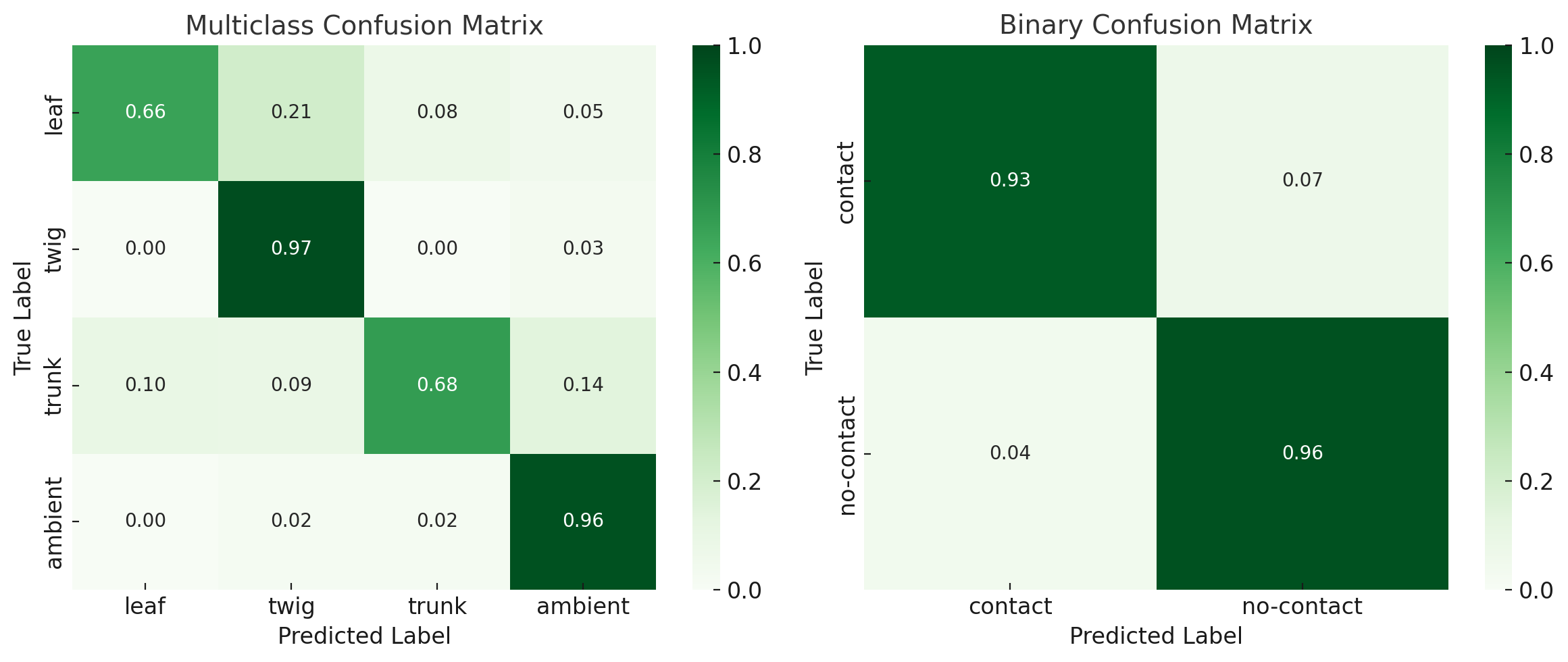}
    \caption{Confusion matrices for (left) four-class contact classification and (right) binary contact detection.  Multiclass F1-score is 0.82 and binary F1-score is 0.94, showing the effectiveness of our method. }
    \label{fig:modality_ablation-robot}
    \vspace{-10pt}
\end{figure}

Our research investigates three fundamental questions concerning the classification of object contact interactions through sensory modalities.

During training on the hand-held probe data, all single modality models, encompassing both audio only and vision-only baselines, achieve high accuracy with the controlled probe recordings. However, the vision-only model, lacking access to audio cues, struggles to differentiate between contact and non-contact events, often misinterpreting ambient interactions as subtle contact types. Figure~\ref{fig:probe_confusion_matrix} compares our method (left) with the vision-only model (right) on probe recordings. While the vision-only model demonstrates frequent cross category errors, our fusion method achieves near perfect classification across all contact types, highlighting that the integration of spectral audio information with visual context effectively mitigates ambiguity in contact categorization.

To answer \textbf{Q1}, we evaluate audio’s standalone capabilities using two complementary pretrained encoders, the Audio Spectrogram Transformer (AST) and the Contrastive Language–Audio Pretraining (CLAP) model—fine-tuned on our probe-collected dataset and tested on out-of-distribution robot recordings. When AST and CLAP are combined, they achieve a multiclass F1 score of 0.53 (Table \ref{tab:multiclass-performance}), demonstrating that audio alone encodes rich vibrotactile cues sufficient to distinguish contact categories. For reference, random guess baseline achieves F1 score of 0.28. Audio-only outperforms the image-only ViT model, which achieves a lower multiclass F1 of 0.35 despite being pretrained, indicating that visual signals alone are less informative for classifying nuanced contact interactions. These results underscore the importance of audio for interpreting subtle, contact-rich events that are often obscured or ambiguous in visual input alone.

In the next phase of our study, for \textbf{Q2}, we investigated multimodal fusion by integrating the audio features derived from the AST and CLAP with a Vision Transformer (ViT) as discussed in Sec~\ref{sec:classification}. The performance of the fused model was compared against its audio-only and image-only counterparts. The audio-only attained an F1 score of 0.91 for binary detection (Table \ref{tab:binary-performance}) and 0.53 for multiclass classification (Table \ref{tab:multiclass-performance}), while the pretrained ViT achieved scores of 0.57 and 0.35, respectively. Notably, the fusion of audio and image data significantly enhanced multiclass performance, resulting in an F1 score of 0.74 while maintaining binary detection capabilities (F1 = 0.90). This finding demonstrates that visual cues, such as shape and texture at contact points, complement audio information, particularly in situations characterized by occlusion or ambiguous auditory signals. For example, while the vision-only model frequently confused leaf and twig classes (Figure \ref{fig:confusion_image}), the fused model successfully distinguished them by leveraging both the visual texture and the subtle differences in contact sound, enabling finer-grained classification.

Finally, to quantify the effects of pretraining in \textbf{Q3}, we compared the performance of each architecture when initialized from large-scale datasets against those trained from scratch. Pretraining notably enhanced the ViT's multiclass F1 score from 0.21 to 0.35 and improved the multiclass performance of the AST and CLAP from 0.46 to 0.53. In the context of the fused model, pretraining yielded the most substantial improvement, increasing the multiclass F1 score from 0.52 (trained from scratch) to 0.74. These results affirm that leveraging pretrained audio and visual architectures provides robust priors that facilitate effective generalization to robot-collected data subject to domain shifts.

In summary, our findings demonstrate that \textbf{(Q1)} audio alone is proficient for detecting and approximately classifying contact interactions, \textbf{(Q2)} the integration of audio and visual information yields the most robust multiclass performance, and \textbf{(Q3)} pretraining on large-scale external datasets is crucial for achieving high accuracy when adapting to novel sensing conditions.

\subsection{Input Features}\label{sec:classification-inputs}

The advantage of adopting a multi-modal approach over a singular modality is that different modalities can compensate for one another. Not only do multiple modalities enhance the overall performance of classification, but they can also provide support when one modality becomes compromised. For example, in our tree classification scenario, if objects obstruct the camera's view, the model can still rely on sound to compensate for its temporary inability to see. It is important to note that this solution is not always guaranteed; the model may still confuse what it is actually interacting with due to the obstruction. However, the additional context provided by sound has empirically improved the model's predictions. Another example is when the image captures multiple contact categories but is unsure which one to choose. In this case, audio data assisted in differentiating between the two options, as highlighted in Figure \ref{fig:confusion_image}.

\begin{figure}[!t]
    \centering
    \includegraphics[width=1\columnwidth]{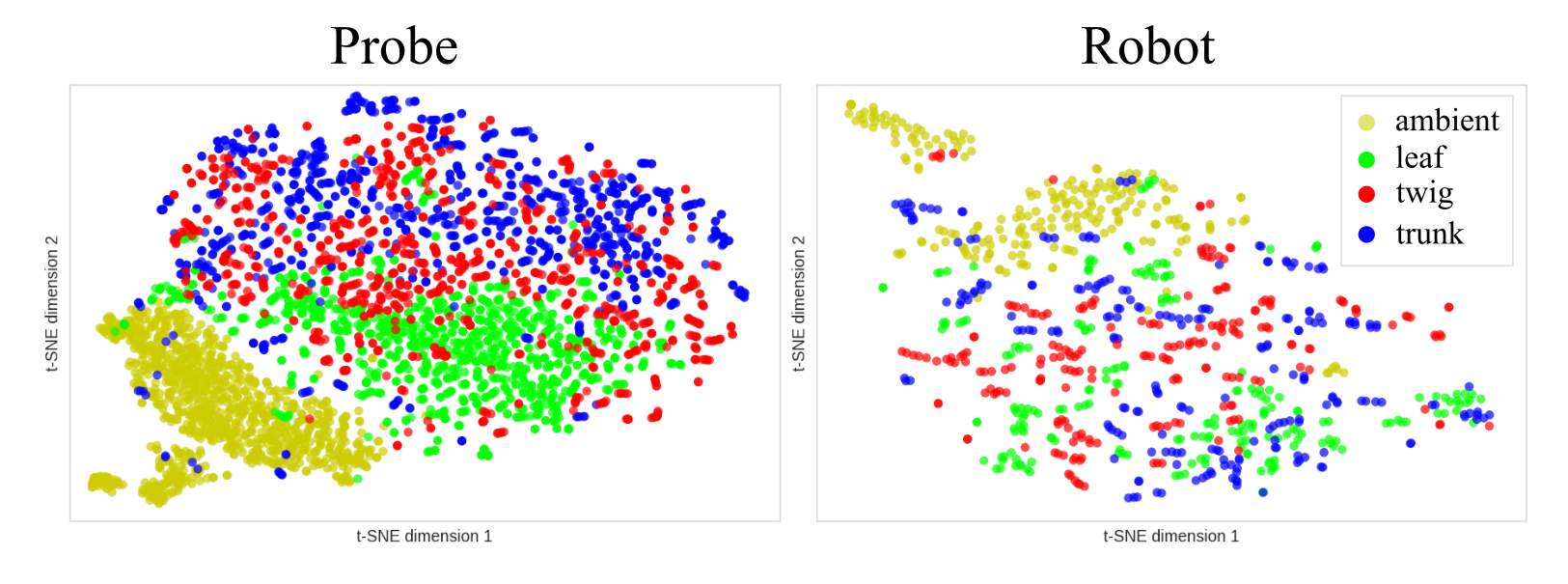}
    \caption{The t-SNE analysis of extracted audio features from our 1-second audio samples reveals interesting insights. The hand held probe data shows significant clustering of categories, while the robot data does not exhibit this pattern. This observation highlights the necessity of implementing a machine learning model to help classify contact interactions.}
    \label{fig:tsne}
    \vspace{-10pt}
\end{figure}

\subsection{Qualitative Evalulation}
From our dataset, we extract audio and calculate scalar values, such as root mean square, MFCCs, zero crossing rate, and others, which we then forward to a t-SNE model using 70\% of the data. We present the clusterings of categories based solely on audio samples. As shown in Figure \ref{fig:tsne}, probe data demonstrates a clear separation between ambient and leaf categories; however, there is some notable overlap between the twig and trunk categories. In contrast, the robot data is not as distinctly clustered, with the exception of the ambient section. Given that the robot data presents challenges in distinguishing between categories, it is imperative that we employ a machine learning model to better learn the nuances of data separation.

\begin{figure}[!t]
    \centering
    \includegraphics[width=1\columnwidth]{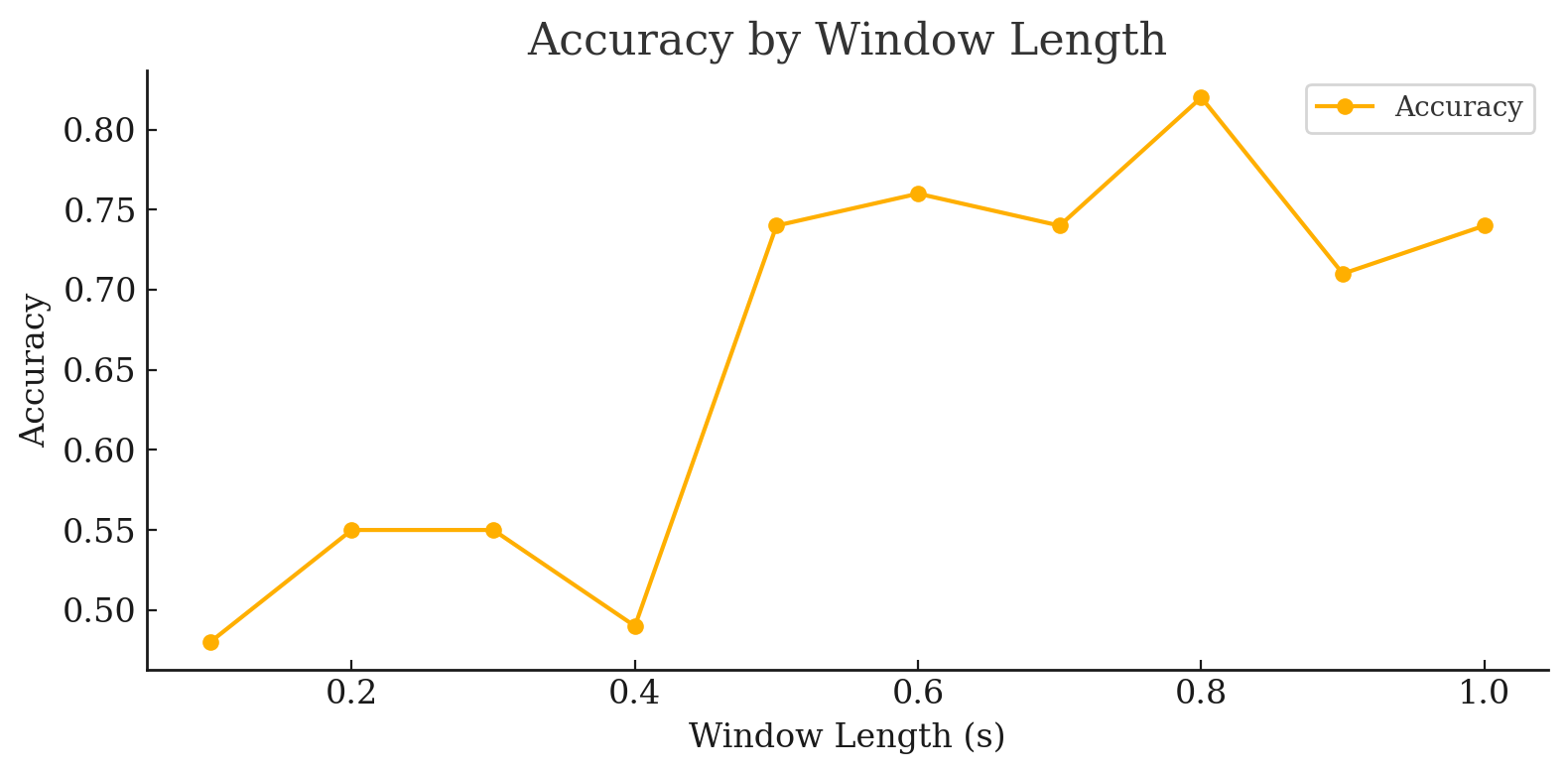}
    \caption{Classification accuracy in relation to audio window length, ranging from 0.1 to 1.0 seconds. Circles at each 0.1-second interval are connected by a bold line, showing that accuracy increases to 0.82 at 0.8 seconds before tapering off. This indicates that the optimal time window for sampling is around 0.8 seconds; any additional information may confuse the model, while shorter windows do not provide enough data.}
    \label{fig:time}
    \vspace{-10pt}
\end{figure}

\subsection{Additional Ablation}
Figure \ref{fig:time} illustrates the relationship between classification accuracy on the multiclass task and audio clip duration, which varies from 0.1 seconds to 1.0 seconds in 0.1-second increments. All tests were conducted using the same number of samples at each window length to ensure a fair comparison. The data reveal a marked sensitivity to the length of the time window employed for analysis. Specifically, very short windows (0.1–0.4 s) yield notably lower and more variable accuracy (0.46–0.55), reflecting insufficient temporal context. Performance then improves sharply at 0.5 s (0.72) and continues rising to a peak of 0.82 at 0.8s. Beyond this optimal duration, accuracy declines to 0.74 at 1.0 s, suggesting that longer windows may introduce redundant or noisy information that diminishes classification performance.
These results indicate that a window of approximately 0.8 seconds strikes the best balance between capturing enough contact-related temporal structure and avoiding extraneous data for achieving high accuracy. Shorter windows, for example 0.3–0.5 seconds, can trade a slight accuracy loss for noticeably lower latency when fast reactions are desirable. %

\section{Conclusion}

We presented a multi-modal classification framework that uses audio and visual inputs to identify contact interactions with tree structures in agricultural settings. By combining vibrotactile sensing from contact microphones with visual cues, our system can classify contact events into semantic categories: leaf, twig, trunk, or ambient. Our experimental results show that the combined use of vibrotactile and visual modalities significantly improves classification accuracy over unimodal baselines. Furthermore, with audio preprocessing and pretraining, our results show model's generalization ability to zero-shot transfer to different hardware embodiments (e.g., hand-held probe to robot probe). Despite these results, our current method relies only on single-frame audio-visual snapshots and does not explicitly model temporal continuity, making it vulnerable to contact transitions or edge cases. Future work will explore temporal filtering and model architecture to handle sequence predictions to improve classification robustness based on prior information. Additionally, while this work focuses on contact classification, future efforts will integrate semantic contact understanding into closed-loop control to inform safe and reactive motion policies during contact-rich manipulation.

 %

\section*{Acknowledgment}
This work was supported by NSF Robust Intelligence 1956163 and NSF/USDA NIFA AIIRA AI Research Institute 2021-67021-35329.

\bibliographystyle{IEEEtran} 
\bibliography{mybib}

\begin{thebibliography}{10}
\providecommand{\url}[1]{#1}
\csname url@samestyle\endcsname
\providecommand{\newblock}{\relax}
\providecommand{\bibinfo}[2]{#2}
\providecommand{\BIBentrySTDinterwordspacing}{\spaceskip=0pt\relax}
\providecommand{\BIBentryALTinterwordstretchfactor}{4}
\providecommand{\BIBentryALTinterwordspacing}{\spaceskip=\fontdimen2\font plus
\BIBentryALTinterwordstretchfactor\fontdimen3\font minus \fontdimen4\font\relax}
\providecommand{\BIBforeignlanguage}[2]{{%
\expandafter\ifx\csname l@#1\endcsname\relax
\typeout{** WARNING: IEEEtran.bst: No hyphenation pattern has been}%
\typeout{** loaded for the language `#1'. Using the pattern for}%
\typeout{** the default language instead.}%
\else
\language=\csname l@#1\endcsname
\fi
#2}}
\providecommand{\BIBdecl}{\relax}
\BIBdecl

\bibitem{roy2021machine}
N.~Roy, I.~Posner, T.~Barfoot, P.~Beaudoin, Y.~Bengio, J.~Bohg, O.~Brock, I.~Depatie, D.~Fox, D.~Koditschek \emph{et~al.}, ``From machine learning to robotics: Challenges and opportunities for embodied intelligence,'' \emph{arXiv preprint arXiv:2110.15245}, 2021.

\bibitem{lee2024sonicboom}
M.~Lee, U.~Yoo, J.~Oh, J.~Ichnowski, G.~Kantor, and O.~Kroemer, ``Sonicboom: Contact localization using array of microphones,'' \emph{arXiv preprint arXiv:2412.09878}, 2024.

\bibitem{hellebrekers2019soft}
T.~Hellebrekers, O.~Kroemer, and C.~Majidi, ``Soft magnetic skin for continuous deformation sensing,'' \emph{Advanced Intelligent Systems}, vol.~1, no.~4, p. 1900025, 2019.

\bibitem{yuan2017gelsight}
W.~Yuan, S.~Dong, and E.~H. Adelson, ``Gelsight: High-resolution robot tactile sensors for estimating geometry and force,'' \emph{Sensors}, vol.~17, no.~12, p. 2762, 2017.

\bibitem{bhirangi2021reskin}
R.~Bhirangi, T.~Hellebrekers, C.~Majidi, and A.~Gupta, ``Reskin:versatile, replaceable, lasting tactile skins,'' in \emph{CoRL}, 2021.

\bibitem{thankaraj2023sounds}
A.~Thankaraj and L.~Pinto, ``That sounds right: Auditory self-supervision for dynamic robot manipulation,'' in \emph{Conference on Robot Learning}.\hskip 1em plus 0.5em minus 0.4em\relax PMLR, 2023, pp. 1036--1049.

\bibitem{liu2024maniwav}
Z.~Liu, C.~Chi, E.~Cousineau, N.~Kuppuswamy, B.~Burchfiel, and S.~Song, ``Maniwav: Learning robot manipulation from in-the-wild audio-visual data,'' in \emph{8th Annual Conference on Robot Learning}, 2024.

\bibitem{kim2023occlusion}
C.~H. Kim and G.~Kantor, ``Occlusion reasoning for skeleton extraction of self-occluded tree canopies,'' in \emph{2023 IEEE International Conference on Robotics and Automation (ICRA)}.\hskip 1em plus 0.5em minus 0.4em\relax IEEE, 2023, pp. 9580--9586.

\bibitem{lagrassa2024task}
A.~LaGrassa, M.~Lee, and O.~Kroemer, ``Task-oriented active learning of model preconditions for inaccurate dynamics models,'' \emph{arXiv}, 2024.

\bibitem{Kantor_visual}
F.~Yandun, A.~Silwal, and G.~Kantor, ``{Visual 3D Reconstruction and Dynamic Simulation of Fruit Trees for Robotic Manipulation},'' in \emph{IEEE/CVF Conference on Computer Vision and Pattern Recognition Workshops}, 2020, pp. 54--55.

\bibitem{sorghum_3d}
H.~Freeman, E.~Schneider, C.~H. Kim, M.~Lee, and G.~Kantor, ``3d reconstruction-based seed counting of sorghum panicles for agricultural inspection,'' in \emph{2023 IEEE International Conference on Robotics and Automation (ICRA)}, 2023, pp. 9594--9600.

\bibitem{jacob2024learning}
J.~Jacob, T.~Bandyopadhyay, J.~Williams, P.~Borges, and F.~Ramos, ``Learning to simulate tree-branch dynamics for manipulation,'' \emph{IEEE Robotics and Automation Letters}, vol.~9, no.~2, pp. 1748--1755, 2024.

\bibitem{kim2023robotic}
C.~H. Kim, M.~Lee, O.~Kroemer, and G.~Kantor, ``Towards robotic tree manipulation: Leveraging graph representations,'' 2023.

\bibitem{jacobgentle}
J.~Jacob, S.~Cai, P.~V.~K. Borges, T.~Bandyopadhyay, and F.~Ramos, ``Gentle manipulation of tree branches: A contact-aware policy learning approach,'' in \emph{8th Annual Conference on Robot Learning}, 2024.

\bibitem{liu2024sonicsense}
J.~Liu and B.~Chen, ``Sonicsense: Object perception from in-hand acoustic vibration,'' \emph{arXiv preprint arXiv:2406.17932}, 2024.

\bibitem{yi2024visual}
X.~Yi, J.~Lee, and N.~Fazeli, ``Visual-auditory extrinsic contact estimation,'' \emph{arXiv preprint arXiv:2409.14608}, 2024.

\bibitem{du2022play}
M.~Du, O.~Y. Lee, S.~Nair, and C.~Finn, ``Play it by ear: Learning skills amidst occlusion through audio-visual imitation learning,'' \emph{arXiv preprint arXiv:2205.14850}, 2022.

\bibitem{clarke2018learning}
S.~Clarke, T.~Rhodes, C.~G. Atkeson, and O.~Kroemer, ``Learning audio feedback for estimating amount and flow of granular material,'' \emph{Proceedings of Machine Learning Research}, vol.~87, 2018.

\bibitem{gandhi2020swoosh}
D.~Gandhi, A.~Gupta, and L.~Pinto, ``Swoosh! rattle! thump!--actions that sound,'' \emph{arXiv preprint arXiv:2007.01851}, 2020.

\bibitem{chen2021boombox}
B.~Chen, M.~Chiquier, H.~Lipson, and C.~Vondrick, ``The boombox: Visual reconstruction from acoustic vibrations,'' \emph{arXiv preprint arXiv:2105.08052}, 2021.

\bibitem{IAM_cutting}
K.~Zhang, M.~Sharma, M.~Veloso, and O.~Kroemer, ``{Leveraging multimodal haptic sensory data for robust cutting},'' in \emph{IEEE-RAS 19th International Conference on Humanoid Robots (Humanoids)}, 2019.

\bibitem{IAM_food}
A.~Sawhney, S.~Lee, K.~Zhang, M.~Veloso, and O.~Kroemer, ``{Playing with food: Learning food item representations through interactive exploration},'' in \emph{International Symposium on Experimental Robotics}, 2020, pp. 309--322.

\bibitem{brock2023_journal}
V.~Wall, G.~Z{\"o}ller, and O.~Brock, ``Passive and active acoustic sensing for soft pneumatic actuators,'' \emph{The International Journal of Robotics Research}, vol.~42, no.~3, pp. 108--122, 2023.

\bibitem{yoo2024poe}
U.~Yoo, Z.~Lopez, J.~Ichnowski, and J.~Oh, ``Poe: Acoustic soft robotic proprioception for omnidirectional end-effectors,'' \emph{arXiv preprint arXiv:2401.09382}, 2024.

\bibitem{SamsungAI_2023sonicfinger}
S.~Rupavatharam, C.~Escobedo, D.~Lee, C.~Prepscius, L.~Jackel, R.~Howard, and V.~Isler, ``Sonicfinger: Pre-touch and contact detection tactile sensor for reactive pregrasping,'' in \emph{2023 IEEE International Conference on Robotics and Automation (ICRA)}.\hskip 1em plus 0.5em minus 0.4em\relax IEEE, 2023, pp. 12\,556--12\,562.

\bibitem{yang2024binding}
F.~Yang, C.~Feng, Z.~Chen, H.~Park, D.~Wang, Y.~Dou, Z.~Zeng, X.~Chen, R.~Gangopadhyay, A.~Owens \emph{et~al.}, ``Binding touch to everything: Learning unified multimodal tactile representations,'' in \emph{Proceedings of the IEEE/CVF Conference on Computer Vision and Pattern Recognition}, 2024, pp. 26\,340--26\,353.

\bibitem{mejia2024hearing}
J.~Mejia, V.~Dean, T.~Hellebrekers, and A.~Gupta, ``Hearing touch: Audio-visual pretraining for contact-rich manipulation,'' in \emph{2024 IEEE International Conference on Robotics and Automation (ICRA)}.\hskip 1em plus 0.5em minus 0.4em\relax IEEE, 2024, pp. 6912--6919.

\bibitem{yu2021lab}
X.~Yu, Z.~Fan, X.~Wang, H.~Wan, P.~Wang, X.~Zeng, and F.~Jia, ``A lab-customized autonomous humanoid apple harvesting robot,'' \emph{Computers \& Electrical Engineering}, vol.~96, p. 107459, 2021.

\bibitem{silwal2022bumblebee}
A.~Silwal, F.~Yandun, A.~K. Nellithimaru, T.~Bates, and G.~Kantor, ``Bumblebee: A path towards fully autonomous robotic vine pruning.'' \emph{Field Robotics}, vol.~2, no.~1, pp. 1661--1696, 2022.

\bibitem{sainburg2020finding}
T.~Sainburg, M.~Thielk, and T.~Q. Gentner, ``Finding, visualizing, and quantifying latent structure across diverse animal vocal repertoires,'' \emph{PLoS computational biology}, vol.~16, no.~10, p. e1008228, 2020.

\bibitem{park2022biomimetic}
K.~Park, H.~Yuk, M.~Yang, J.~Cho, H.~Lee, and J.~Kim, ``A biomimetic elastomeric robot skin using electrical impedance and acoustic tomography for tactile sensing,'' \emph{Science Robotics}, vol.~7, no.~67, p. eabm7187, 2022.

\bibitem{gong2021ast}
Y.~Gong, Y.-A. Chung, and J.~Glass, ``Ast: Audio spectrogram transformer,'' \emph{arXiv preprint arXiv:2104.01778}, 2021.

\bibitem{wu2023large}
Y.~Wu, K.~Chen, T.~Zhang, Y.~Hui, T.~Berg-Kirkpatrick, and S.~Dubnov, ``Large-scale contrastive language-audio pretraining with feature fusion and keyword-to-caption augmentation,'' in \emph{ICASSP 2023-2023 IEEE International Conference on Acoustics, Speech and Signal Processing (ICASSP)}.\hskip 1em plus 0.5em minus 0.4em\relax IEEE, 2023, pp. 1--5.

\end{thebibliography}

\end{document}